# Machine learning for discovering laws of nature

Lizhi Xin[1], Kevin Xin[2], Houwen Xin[3, *]

[1] Hefei, P. R. China
[2] Chicago, USA
[3] Department of Chemical physics USTC, Hefei, Anhui, P. R. China
[*] hxin@ustc.edu.cn

## ABSTRACT

Based on Darwin's natural selection, we developed "machine scientists" to discover the laws of nature by learning from raw data. "Machine scientists" construct physical theories by applying a logic tree (state Decision Tree) and a value tree (observation Function Tree); the logical tree determines the state of the entity, and the value tree determines the absolute value between the two observations of the entity. A logic Tree and a value tree together can reconstruct an entity's trajectory and make predictions about its future outcomes. Our proposed algorithmic model has an emphasis on machine learning - where "machine scientists" builds up its experience by being rewarded or punished for each decision they make - eventually leading to rediscovering Newton's equation (classical physics) and the Born's rule (quantum mechanics).

## Introduction

A fundamental physics theory includes three core elements: 1) to describe the observed experiment data; 2) to predict the future outcome; 3) to understand natural phenomena. To this day, the mainstream way of scientific discovery is still: 1) to describe experimental observations in rigorous mathematical structure (differential equation); 2) to solve the differential equation under the constraints of the initial conditions and boundary conditions, from which the change from the initial state to the final state of the particle can be obtained, and the future state of the particle can be further predicted; 3) to explain natural phenomena by defining the most basic concepts about nature, such as force (classical mechanics), field (classical electromagnetism), wave functions (quantum mechanics), entropy (classical statistical mechanics), and so on. So far, under the framework of rigorous mathematical models, the laws of physics are beautifully described by calculus equations.

   Mathematical models are good for simple closed systems and have been a great success; examples include Newton's equations (classical mechanics), Maxwell's equations (classical electromagnetism), and Schrödinger's equations (quantum mechanics). However, due to the complexity of open systems, such solutions usually do not work. Recent research has proposed the development of "machine scientists" to distill the laws of physics from raw data. Roger Guimera and Marta Sales-Pardo [1] developed a symbolic regression algorithm called the Bayesian machine scientist; Patrick Langley [2] developed BACON to rediscover Kepler's third law; Lipson and Michael Schmidt [3] applied genetic programming to develop an algorithm called Eureqa which successfully recover equations describing the motion of one pendulum hanging from another; Steven Brunton, Joshua Proctor and Nathan Kutz [4] developed a "machine scientist" algorithm by applying sparse regression; Miles Cranmer et al. [5] has developed an symbolic regression algorithm directly inferring Newton's law of gravitation; Max Tegmark and Silviu-Marian Udrescu [6] has developed a "machine scientist" called "AI Feynman" to rediscover 100 equations from the Feynman Lectures on Physics; Cristina Cornelio et al. [7] has developed a "machine scientist" called "AI-Descartes" for scientific discovery.; Krenn et al. [8] has written a survey article about scientific understanding with artificial intelligence.

   The "machine scientists" mentioned above are all based on rigorous mathematical models to describe and explain the laws of physics, that is, to find mathematical equations to describe regularities by learning raw data. We believe that the laws of nature aren't simply defined by calculus (rigorous mathematical model), but by evolutionary algorithms (dynamic rules); A mathematical model is just a simple representation of dynamic rules.

We propose a computational model based on Darwin's natural selection to discover physical laws. We do not model with the usual function (classical physics) or wave function (quantum physics), but a state Decision Tree (logic) and an observation Function Tree (value); the state decision tree, together with the observation function tree, can reconstruct the trajectories of entities (macro and micro) and predict their future trajectories. In this paper, "machine scientists" are developed to construct physical theories. "Machine scientists" discover laws of physics by learning the raw data with genetic programming. Our model emphasizes machine learning, where "machine scientist" builds up its experience of the outside world by being rewarded or punished for each decision it makes, and eventually rediscovers Newton's equation (classical physics) and Born's rule (quantum mechanics).

## Machine learning algorithm

Both an ice hockey puck (macroscopic) and an electron (microscopic) can be considered a generic entity; this entity that dynamically changes over time in terms of time series consisting of states and observations can be uniformly defined as a finite set points as in (1):

$$\{(q_k, x_k)\} \; k = 1, \cdots, N \tag{1}$$

Where $q_k$ denotes dynamic state of the entity, $x_k$ denotes the observed value of the entity, and the data sequence $\{x_k, k = 1, \cdots, N\}$ defines the "trajectory" of the entity; if $x_k \geq x_{k-1}$, the state of the entity is 0 ($q_k = 0$); if $x_k < x_{k-1}$, the state of the entity is 1 ($q_k = 1$).

"Machine scientists" can learn from time series $\{(q_k, x_k)\}$ to build up experience, and based on the accumulated experience (knowledge), construct satisfactory physical theories to understand nature. The question now becomes: Can "machine scientists" find a satisfactory theory? In other words, given a time series $\{(q_k, x_k)\}$ as input, can a "machine scientist" construct a theory $t_k$ capable of: 1) generate results that match the observations of the input; 2) predict the future outcome; 3) understand nature.

Our solution is to apply quantum-like machine learning to discover the laws of physics. A quantum-like machine learning algorithm can be expressed in terms of a two-part "trees": The first part is the observation Function Tree (value tree) that describes the observed value of the entity and the other part is the state Decision Tree (logical tree), which applies "yes or no" logic to determine the state of the entity[9]. Together, observation Function Tree and state Decision Tree will reconstruct the "trajectory" of the entity.

1. Observation Function Tree (xFT): xFT is used to calculate the absolute "distance" between two points of an entity ($|x'_k - x'_{k-1}|$).

2. State Decision Tree (qDT): qDT determines the state of the entity $q'_k$ and calculates the theoretical value of the entity $x'_k$.

As shown in Figure 1, a satisfactory theory $t_k$ (2a) can: 1) generate the results to match measured outcomes (2b); 2) predict the outcomes in the future (2c); 3) understand nature.

$$\{(q_k, x_k)\} \xrightarrow{\text{input}} t_k(xFT, qDT) \xrightarrow{\text{output}} \{(q'_k, x'_k)\} \tag{2a}$$

$$1) \quad q'_k = q_k \text{ and } x'_k = x_k \tag{2b}$$

$$2) \quad q'_n = q_n \text{ and } x'_n = x_n \tag{2c}$$

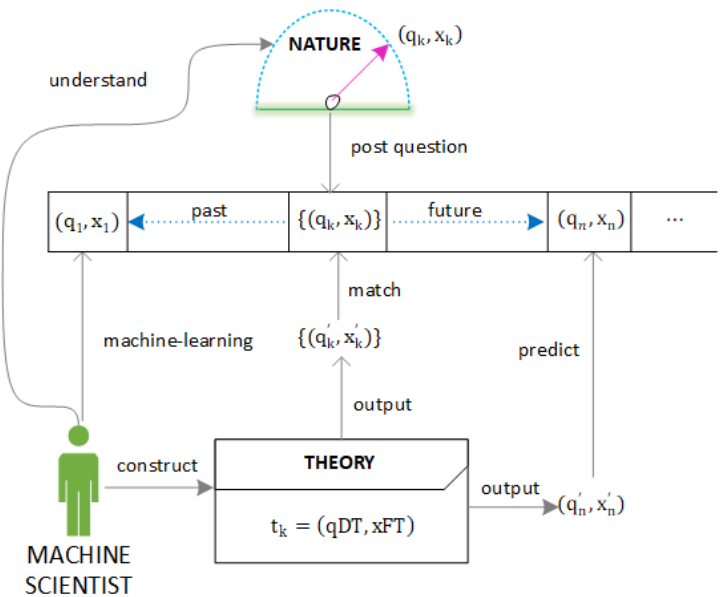

**Figure 1** Machine scientist-theory-nature.

We propose the use of genetic programming (GP) [10-13] to discover "satisfactory" physical theories. In his paper (*Genetic Programming: A Paradigm for Genetically Breeding Populations of Computer Programs to Solve Problems*, Stanford University Computer Science Department technical report STAN-CS-90-1314.), John Koza started the abstract off with:

*"Many seemingly different problems in artificial intelligence, symbolic processing, and machine learning can be viewed as requiring discovery of a computer program that produces some desired output for particular inputs. When viewed in this way, the process of solving these problems becomes equivalent to searching a space of possible computer programs for a most fit individual computer program. The new "genetic programming" paradigm described herein provides a way to search for this most fit individual computer program. In this new "genetic programming" paradigm, populations of computer programs are genetically bred using the Darwinian principle of survival of the fittest and using a genetic crossover (recombination) operator appropriate for genetically mating computer programs. In this paper, the process of formulating and solving problems using this new paradigm is illustrated using examples from various areas."*

Koza goes on to state in the introduction:

*"Depending on the terminology of the particular field involved, the "computer program" may be called a robotic action plan, an optimal control strategy, a decision tree, an econometric model, a set of state transition equations, the transfer function, a game-playing strategy, or, perhaps more generically, a "composition of functions." Similarly, the "inputs" to the "computer program" may be called sensor values, state variables, independent variables, attributes, or, perhaps merely, "arguments to a function." However, regardless of different terminology used, the underlying common problem is discovery of a computer program that produces some desired output when presented with particular inputs. The purpose of this paper is to show how to reformulate these seemingly different problems into a common form (i.e. a problem requiring discovery of a computer program) and, then, to describe a single, unified approach for solving problems formulated in this common form."*

The idea and steps of GP are simple: 1) randomly generate 300~500 theories; 2) study historical data to obtain fitness for each theory; 3) the "satisfactory" theory is obtained through Darwinian principle of survival of the fittest (crossover, mutation and selection) after about 100 generations of evolution.

**Genetic programming algorithm**
*Input*:

- Time series $\{(q_k, x_k), k = 0, \cdots, N\}$ (each sample consists of an entity's state and value).
- Setting
  1) Operation set F
  2) Data set T
  3) Crossover probability = 70%; Mutation probability = 5%.

*Initialization*:
- Population: randomly create 300-500 individuals.

*Evolution*:
- for i = 0 to 100 generations
  a) Calculate fitness for each individual based on historical data set.
  b) According to the quality of fitness:
     i. Selection: selecting parents.
     ii. Crossover: generate a new offspring using the roulette algorithm based on crossover probability.
     iii. Mutation: randomly modify parent based on mutation probability.

*Output*:
- An individual of the best fitness.

**observation Function Tree (xFT)**

An xFT consists of a traditional GP function tree. The final form of the xFT is represented as a function, and the output of the xFT is a numeric value. For xFT the operation set F and data set T are as follows

1) Operation set $F = \{+, -, \times, \div, \sin, \cos, \log, \exp\}$

2) Data set $T = \{k, z_1, z_2, \cdots, z_m\}$

Where k denotes kth element of a data sequence, $z_m$ denotes a variable, data set $T = \{z_1, z_2, \cdots, z_m\}$ depends on the problem space, such as velocity ($z_m = v$) of a particle for classical physics. xFT is a function that consists of operation set F and data set T.

$$xFT = f(F, T) \tag{3}$$

We define the "absolute distance" between two observations of an entity as follows:

$$d_{k,k-1} = |x_k - x_{k-1}| \tag{4}$$

A "machine scientist" can calculate the "absolute distance" between two observations of an entity by xFT:

$$d'_{k,k-1} = f(F, \{k, z_1, z_2, \cdots, z_m\}) - f(F, \{k-1, z_1, z_2, \cdots, z_m\}) \tag{5}$$

Now we can define fitness function for xFT as follows:

$$xFT_{fitness} = -\sum_{k=1}^{N}(d'_{k,k-1} - d_{k,k-1})^2 \tag{6}$$

Where $d'_{k,k-1}$ is the theoretical value which is calculated by xFT (5), and $d_{k,k-1}$ is the actual observed value (4). What the fitness function essentially is a particular type of function that is used to summarize, as a single figure of merit, how close a given design solution is to achieving the set aims. Fitness functions are used in genetic programming to guide simulations towards optimal design solutions. In order to reach the optimal solution, the GP algorithm implements a continuous evolution process through selection, crossover, and mutation. The goal of continuous evolution is to find a satisfactory xFT that makes $d'_{k,k-1}$ as close to $d_{k,k-1}$ as possible.

**state Decision Tree (qDT)**

A qDT consists of a GP matrix tree which is constructed from 8 basic quantum gates (matrices). The final form of qDT is represented as a matrix. The output from this matrix is a vector (for example an action: $\begin{bmatrix}1\\0\end{bmatrix}$ represents that the "machine scientist" believed that atoms do not decay or $\begin{bmatrix}0\\1\end{bmatrix}$ represents that the "machine scientist" believed that atoms will decay). The purpose of qDT is to simulate the decision-making process of "machine scientist" under uncertainty. For example, an atom may have two states, non-decay or decay, represented by $q_1$ and $q_2$ respectively; "machine scientists" have two possible actions, believing that atoms do not decay, or believing that atoms do decay, represented by $a_1$ and $a_2$, respectively. The uncertain state of whether an atom decays can be expressed by the superposition of all possible states:

$$|\psi\rangle = c_1|q_1\rangle + c_2|q_2\rangle \tag{7}$$

Where $|q_1\rangle$ indicates a state in which the atom does not decay and $|q_2\rangle$ indicates a state in which the atom decays. $|c_1|^2$ is the frequency at which the atom does not decay; $|c_2|^2$ is the frequency at which the atom decays.

We hypothesize that when a "machine scientist" is hesitant to choose an action, it's decision can be represented by the superposition of all possible actions:

$$|\phi\rangle = \mu_1|a_1\rangle + \mu_2|a_2\rangle \tag{8}$$

Where $|a_1\rangle$ indicates that "machine scientists" believe that atoms do not decay, and $|a_2\rangle$ indicates that "machine scientists" believe that atoms will decay; $p_1 = |\mu_1|^2$ represents the probability that the "machine scientist" chooses the action $|a_1\rangle$, and $p_2 = |\mu_2|^2$ represents the probability that the "machine scientist" chooses the action $|a_2\rangle$.

"Machine scientists" have incomplete information about nature before making decisions, so they cannot accurately predict whether the next state of nature will be $|q_1\rangle$ or $|q_2\rangle$, which forces "machine scientists" to "bet". Before a "machine scientist" makes a decision, its decision-making state is in a pure state, a superposition state that can decide whether to act $|a_1\rangle$ and $|a_2\rangle$ at the same time. This pure state [14-16] is when the actions of $|a_1\rangle$ and $|a_2\rangle$ are superposed in its mind. When the "machine scientist" makes the final decision, it's decision-making state changes from pure to mixed, that is, it chooses an action from the available actions with a probability, taking action $a_1$ with probability $p_1$ and acting $a_2$ with probability $p_2$.

Decision process: $\rho = |\phi\rangle\langle\phi| \xrightarrow{\text{decision}} \rho' = p_1|a_1\rangle\langle a_1| + p_2|a_2\rangle\langle a_2|$ (9)

It is expressed in matrix form as follows:

$$\rho = \begin{bmatrix}\rho_{11} & \rho_{12}\\ \rho_{21} & \rho_{22}\end{bmatrix} \xrightarrow{\text{diagonalization}} \begin{bmatrix}\lambda_1 & 0\\ 0 & \lambda_2\end{bmatrix} \xrightarrow{\text{normalization}} \rho' = \begin{bmatrix}p_1 & 0\\ 0 & p_2\end{bmatrix} = p_1|a_1\rangle\langle a_1| + p_2|a_2\rangle\langle a_2| \tag{10a}$$

$$|a_1\rangle = \begin{bmatrix}1\\0\end{bmatrix}, |a_2\rangle = \begin{bmatrix}0\\1\end{bmatrix}; |a_1\rangle\langle a_1| = \begin{bmatrix}1 & 0\\0 & 0\end{bmatrix}, |a_2\rangle\langle a_2| = \begin{bmatrix}0 & 0\\0 & 1\end{bmatrix} \tag{10b}$$

The pure state $\rho$ can be approximately constructed from eight basic quantum gates. For qDT the operation set and data set are as follows:

1) Operation set $F = \{+, \times, //\}$

2) Data set $T = \{H, X, Y, Z, S, D, T, I\}$ [17]

- $\begin{cases} H = \frac{1}{\sqrt{2}}\begin{bmatrix}1 & 1\\1 & -1\end{bmatrix} \ X = \begin{bmatrix}0 & 1\\1 & 0\end{bmatrix} \ Y = \begin{bmatrix}0 & -i\\i & 0\end{bmatrix} \ Z = \begin{bmatrix}1 & 0\\0 & -1\end{bmatrix} \\ S = \begin{bmatrix}1 & 0\\0 & i\end{bmatrix} \ D = \begin{bmatrix}0 & 1\\-1 & 0\end{bmatrix} \ T = \begin{bmatrix}1 & 0\\0 & e^{i\pi/4}\end{bmatrix} \ I = \begin{bmatrix}1 & 0\\0 & 1\end{bmatrix} \end{cases}$

Where + means that two matrices are added, × means that two matrices are multiplied, and // means that one is randomly selected from the two branches of the node. H, X, Y, Z, S, D, T, I are 8 fundamental quantum gates (2X2 matrix). qDT is a state decision tree consisting of operation set F and dataset T, which determines the decision state of the "machine scientist" and calculates the observations of the entity at different times. In other words, with qDT, the "machine scientist" can calculate the state of the entity $q'_k$ and the observation value $x'_k$ ($d'_{k,k-1}$ can be calculated by xFT formula (5)):

$$q'_k = qDT(F, T) = \begin{cases} 0, & \text{"machine scientist" believes entity's state is 0 (probability is } p_1) \\ 1, & \text{"machine scientist" believes entity's state is 1 (probability is } p_2) \end{cases} \quad (11a).$$

$$x'_k = \begin{cases} x'_{k-1} + d'_{k,k-1}, & \text{if } q'_k = 0 \\ x'_{k-1} - d'_{k,k-1}, & \text{if } q'_k = 1 \end{cases} \quad (11b)$$

The next step is to find a way to optimize qDT with a group of satisfactory strategies to guide "machine scientist's" decisions. To optimize anything, there needs to be: 1) a selection of a good evaluation function and 2) how to acquire an optimal solution. The "machine scientist" will try to maximize its expected value when making any decisions. Thus, we need to evaluate how "fit" the result (profit or deficit) of the its decision, which can be done by using expected value as a fitness function to optimize the qDT (equations 12 and 13). The whole idea of having GP go through an iterative evolution loop is to find a satisfactory qDT by means of learning historical data to obtain the most optimal solution. The learning rules are as follows:

1) If the natural state is $q_1$

   i. If the "machine scientist" bets the natural state is $q_1$, the "machine scientist" profits $(x = d_{k,k-1})$;

   ii. If the "machine scientist" bets the natural state is $q_2$, the "machine scientist" deficits $(x = -d_{t,t-1})$;.

2) If the natural state is $q_2$

   i. If the "machine scientist" bets the natural state is $q_2$, the "machine scientist" profits $(x = d_{k,k-1})$;

   ii. If the "machine scientist" bets the natural state is $q_1$, the "machine scientist" deficits $(x = -d_{t,t-1})$;.

The expected value of the k-th "bet" of the "machine scientist" is:

$$EV_k = \begin{cases} = p_1 d_{k,k-1}, & \text{the state is 0 and the "machine scientist" believes the state is 0 with probability } p_1 \\ = -p_2 d_{k,k-1}, & \text{the state is 0 and the machine scientist" believes the state is 1 with probability } p_2 \\ = -p_1 d_{k,k-1}, & \text{the state is 1 and the machine scientist" believes the state is 0 with probability } p_1 \\ = p_2 d_{k,k-1}, & \text{the state is 1 and the machine scientist" believes the state is 1 with probability } p_2 \end{cases} \quad (12)$$

Now we can define fitness function for qFT as follows:

$$qDT_{fitness} = \sum_{k=1}^{N} EV_k \quad (13)$$

$qDT_{fitness}$ maximizes "machine scientist" expectations $(max_{qDT}(\sum_{k=1}^{N} EV_k))$; $xFT_{fitness}$ applies negative feedback to make $\left(d'_{k,k-1} \xrightarrow{equal} d_{k,k-1}\right)$. qDT together with xFT will reconstruct the "trajectory" of the entity $\{(q'_k, x'_k)\} \xrightarrow{equal} \{(q_k, x_k)\}$, and make a prediction about the future outcomes as follows:

$$d'_{k+1,k} = xFT(F, \{k+1, z_1, z_2, \cdots, z_m\}) - xFT(F, \{k, z_1, z_2, \cdots, z_m\}) \quad (14)$$

$$q'_{k+1} = qDT(\{+,\times,//\},\{H,X,Y,Z,S,D,T,I\})$$
$$= \begin{cases} 0, & \text{"machine scientists" believe with probability } p_1 \text{ that the state of entity is } q_1 \\ 1, & \text{"machine scientists" believe with probability } p_2 \text{ that the state of entity is } q_2 \end{cases} \quad (15)$$

$$x'_{k+1} = \begin{cases} x'_k + d'_{k+1,k}, & \text{if } q'_{k+1} = 0) \\ x'_k - d'_{k+1,k}, & \text{if } q'_{k+1} = 1) \end{cases} \quad (16)$$

## Results

### Newton's equation (classical mechanics)

| $q_t$ | $x_t$ |
|---|---|
| 0 | 0 |
| 0 | 7 |
| 0 | 20 |
| 0 | 39 |
| 0 | 64 |
| 0 | 95 |
| 0 | 132 |
| 0 | 175 |
| 0 | 224 |
| 0 | 279 |
| 0 | 340 |
| 0 | 407 |
| 0 | 480 |
| 0 | 559 |
| 0 | 644 |
| 0 | 735 |
| 0 | 832 |
| 0 | 935 |
| 0 | 1044 |
| 0 | 1159 |

**Table 1** the time series of ice hockey puck

$$x = vt + \frac{1}{2}at^2 \quad (17)$$

The first column in Table 1 indicates the state of the puck and the second column indicates the position of the puck; because the position of the puck always increases, so the state of the puck is always 0. The time series of the ice hockey puck can be expressed as in (18). The time series of the positions of the puck in Table 1 is generated by Newton's equation (17). Where x denotes the position of the puck, t denotes the time, v=4 denotes the initial velocity of the puck and a=6 denotes the acceleration of the puck.

$$\{(q_t, x_t)\} \, t = 1,\cdots,20 \quad (18)$$

### xFT for Ice hockey puck

The operation set F and dataset T of the xFT of the puck are as follows:

1) $F = \{+, -, \times, \div\}$

2) $T = \{t, v, a, o, h\}$

Where t denotes time; v denotes velocity; a denotes acceleration; o denotes constant 1; h denotes constant 0.5. Now "machine scientists" can calculate the absolute distance between two positions of the puck via xFT (19):

$$d'_{t,t-1} = f(F, \{t, v, a, o, h\}) - f(F, \{t-1, v, a, o, h\}) \tag{19}$$

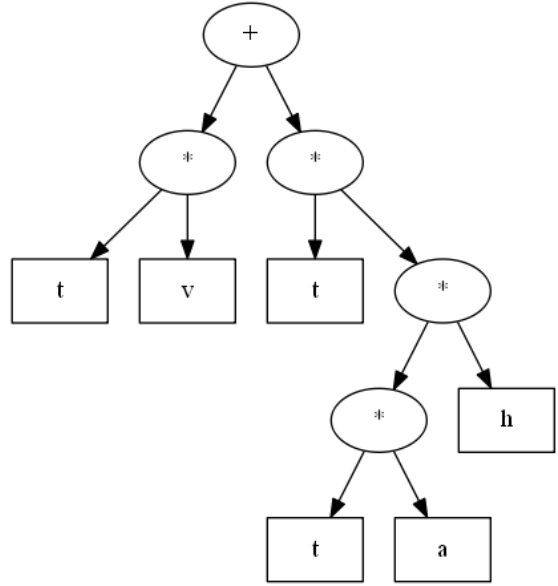

**Figure 2** xFT of ice hockey puck.

By applying the fitness function (6), "machine scientists" evolved an xFT by continuously learning from time series to simulate the continuous changing position of the puck as shown in Figure 2. The corresponding xFT function of Figure 2 is expressed as follows:

$$xFT_{tree}(t) = vt + \frac{1}{2}at^2 \tag{20a}$$

$$d'_{t,t-1} = xFT_{tree}(t) - xFT_{tree}(t-1) = \left(vt + \frac{1}{2}at^2\right) - \left(v(t-1) + \frac{1}{2}a(t-1)^2\right) = v + at - \frac{1}{2}a \tag{20b}$$

**qDT for Ice hockey puck**

qDT determines the state of the puck (21a) and calculates the position of the puck at different times (21b, $d'_{t,t-1}$ can be calculated by (20b)):.

$$q'_t = qDT(F, T) = \begin{cases} 0, & \text{"Machine scientists" believe with a probability } p_1 \text{ that the state of the puck is 0)} \\ 1, & \text{"Machine scientists" believe with a probability } p_2 \text{ that the state of the puck is 1)} \end{cases} \tag{21a}$$

$$x'_t = \begin{cases} x'_{t-1} + d'_{t,t-1}, & \text{if } q'_t = 0 \\ x'_{t-1} - d'_{t,t-1}, & \text{if } q'_t = 1 \end{cases} \tag{21b}$$

As shown in Figure 3, we can obtain a "satisfactory" $qDT_{tree}$ (22a) by using the expected value as a fitness function (13). For this $qDT_{tree}$, the "machine scientist" has only one strategy $S_1$ (22b), i.e. the "machine scientist" is 100% sure that the state of the puck is always 0 (22c). Figure 4 shows that the "machine scientist" reconstructed the trajectory of the puck 100% accurately according to the equation (22d) and will predict the trajectory of the future movement of the puck with 100% accuracy. In other words, without applying differential equations at all, let alone the concept of force, the "machine scientist" independently "discovered" Newton's equation (17).

$$qDT_{tree} = (Y + I) \tag{22a}$$

$$\bullet \quad S_1 = (Y + I) \xrightarrow{\text{diagonalization and normalization}} |a_1\rangle\langle a_1| \tag{22b}$$

$$q'_t = 0 \tag{22c}$$

$$x'_t = x'_{t-1} + d'_{t,t-1} = x'_{t-1} + v + at - \frac{1}{2}a \tag{22d}$$

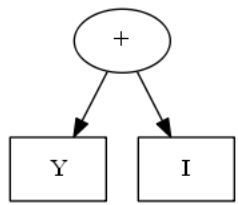

**Figure 3** qDT of ice hockey puck.

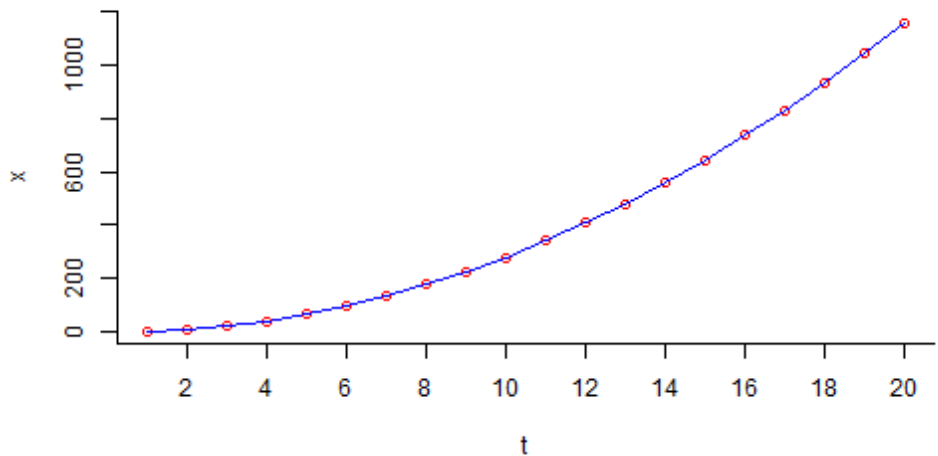

**Figure 4** path of a puck (blue line is the observed one, and the red dots are the computed ones).

### Born's rule (quantum mechanics) [18-32]

In 1935, Erwin Schrödinger created his famous thought experiment involving a cat that is simultaneously both dead and alive to show how absurd quantum physics is.

"A cat is penned up in a steel chamber, along with the following device (which must be secured against direct interference by the cat): in a Geiger counter there is a tiny bit of radioactive substance, so small, that perhaps in the course of the hour one of the atoms decays, but also, with equal probability, perhaps none; if it happens, the counter tube discharges and through a relay releases a hammer which shatters a small flask of hydrocyanic acid. If one has left this entire system to itself for an hour, one would say that the cat still lives if meanwhile no atom has decayed. The psi-function of the entire system would express this by having in it the living and dead cat (pardon the expression) mixed or smeared out in equal parts."

By entangling an atom from the microscopic world with a cat from the macroscopic world, Schrödinger asked a question that is difficult to answer with the Copenhagen Interpretation of quantum mechanics [1-3]: where is the sharp boundary between the quantum world and the classical world? Multiple quantum interpretations spawned from this cat, while the debate regarding it still rages on. A cat in the box will certainly not be in a superposition of dead and alive. According to the quantum measurement theory, observers can only open the box to determine whether the cat is dead or alive, and if the box is not opened for measurement, the observer can only guess probabilistically with the Born's rule, whether the cat is dead or alive.

We will develop a "machine scientist" to describe and explain the Schrödinger cat paradox and the "machine scientist" will build experience (knowledge) by rewarding or punishing every decision it makes, and ultimately "rediscover" the Born's rule through iterative machine learning. Figure 5 shows a modified Schrödinger cat thought experiment designed to allow the "machine scientist" to play games with nature.

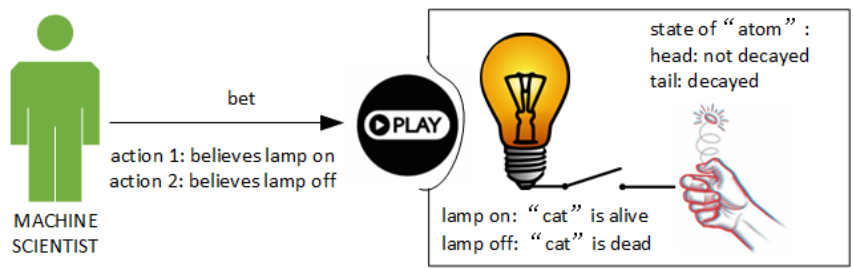

**Figure 5 Modified Schrödinger's cat thought experiment.**

| $q_t$ | $x_t$ |
|---|---|
| 1 | −1 |
| 0 | 0 |
| 0 | 1 |
| 1 | 0 |
| 0 | 1 |
| 1 | 0 |
| 1 | −1 |
| 0 | 0 |
| 0 | 1 |
| 0 | 2 |
| 0 | 3 |
| 1 | 2 |
| 1 | 1 |
| 0 | 2 |
| 1 | 1 |
| 1 | 0 |
| 0 | 1 |
| 1 | 0 |
| 1 | −1 |
| 0 | 0 |

**Table 2** the time series of modified Schrödinger's cat thought experiment.

The rules of the game are as follows:

1) A digital "coin" will be thrown within an hour, if head is up, nothing happens, the lamp is still on; if the tail is up, then a switch will be closed to turn the lamp off

2) If the lamp is on and the "machine scientist" bets the lamp is on, the "machine scientist" wins the game, otherwise the "machine scientist" loses it; if the lamp is off and the "machine scientist" bets the lamp is off, the "machine scientist" wins the game; otherwise the "machine scientist" loses it.

The time series $\{(q_t, x_t)\}$ of modified Schrödinger's cat thought experiment can be generated by the algorithm (23). The algorithm will randomly generate 20 results as shown in Table 2, the first column indicates the cat's state $q_t$ (0 means "cat" is alive, 1 means "cat" is dead), and the second column indicates the observations $x_t$. The data generation algorithm is simple: if the "digital coin" is 0 ($q_t = 0$: indicates that the atom has not decayed), the observation $x_t$ increase by 1; If the "digital coin" is 1 ($q_t = 1$: indicates that the atom decays), the observation $x_t$ decrease by 1.

$$x_{t=0} = 0;\ x_{t+1} = x_t + \begin{cases} 1, & \text{If the "digital coin" is 0} \\ -1, & \text{If the "Digital Coin" is 1} \end{cases} \quad (23)$$

**xFT for Schrödinger's cat**

The operation set F and dataset T of the xFT of the Schrödinger's cat are as follows:

1) $F = \{+, -, \times, \div\}$

2) $T = \{t, d, av, h, l\}$

Where t denotes the tth observation; d denotes average absolute distance between two points; av denotes average value; h denotes highest value; l denotes lowest value. Now "machine scientists" can calculate the absolute distance between two observation points of the Schrödinger's cat via xFT (24):

$$d'_{t,t-1} = f(F, \{t, d, av, h, l\}) - f(F, \{t-1, t, d, av, h, l\}) \quad (24)$$

By applying the fitness function (6), "machine scientists" evolved an xFT (25) by continuously learning from time series as shown in Figure 6. Although the state of Schrödinger's cat changes randomly (the cat may be alive or dead), the absolute "distance" between the two observations of Schrödinger's cat is always 1 (26). It is clear that "machine scientist" has found the satisfactory xFT by studying time series.

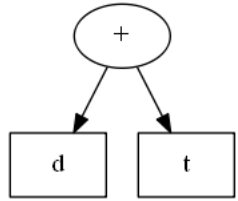

**Figure 6** xFT of Schrödinger's cat.

$$\text{xFT}_{\text{tree}}(t) = d + t \quad (25)$$

$$d'_{t,t-1} = \text{xFT}_{\text{tree}}(t) - \text{xFT}_{\text{tree}}(t-1) = (d+t) - (d+t-1) = 1 \quad (26)$$

**qDT for Schrödinger's cat**

The state of Schrödinger's cat can be represented by the superposition of all possible states:

$$|\psi\rangle = c_1|q_1\rangle + c_2|q_2\rangle \tag{27}$$

Where $|q_1\rangle$ indicates that Schrödinger's cat is alive and $|q_2\rangle$ indicates that Schrödinger's cat is in a dead state. $|c_1|^2$ Indicates the frequency of Schrödinger's cat being alive, and $|c_2|^2$ indicates the objective frequency of Schrödinger's cat being dead.

Without opening the box, the "machine scientist" won't know if the cat is alive or dead, so the "machine scientist" can only subjectively choose an "action," that is, guessing whether the cat is alive or the cat is dead. According to the tripartite model, one can have three attitudes towards the event E: believe E will happen, disbelieve E will happen, and withhold belief and haven't actually decided whether E will happen or not. We hypothesize that when a "machine scientist" hesitates to make a decision, its state of mind can be represented by the superposition of all possible actions:

$$|\phi\rangle = \mu_1|a_1\rangle + \mu_2|a_2\rangle \tag{28}$$

Where $|a_1\rangle$ indicates that the "machine scientist" believes that the cat is alive, $|a_2\rangle$ means that the "machine scientist" believes that the cat is dead, $p_1 = |\mu_1|^2$ indicates the subjective probability that the "machine scientist" chooses the action $|a_1\rangle$, and $p_2 = |\mu_2|^2$ indicates the subjective probability that the "machine scientist" chooses the action $|a_2\rangle$. The outcome of the "machine scientist" decision ("bet") depends on the state of the cat and the action taken by the "machine scientist". In other words, different actions taken by "machine scientists" will lead to different results (guessing right or wrong), and if they guess correctly, the "machine scientist" will be rewarded, and if they guess wrong, they will be punished. The "machine scientist" builds its experience (knowledge) precisely by constantly learning the outcome of the "bet" (reward or punishment).

The information about Schrödinger's cat obtained by the "machine scientist" before placing the bet is incomplete. Before a "machine scientist" makes a decision, its decision-making state is in a pure state, a superposition state that allows it to decide whether to simultaneously choose to believe that the cat is both alive and dead. But in reality, "machine scientists" can't bet on cats both alive and dead at the same time. This pure state is when the cat is both alive and dead superimposed in the "brain" of the "machine scientist". When the "machine scientist" makes the final decision, the "machine scientist's" decision-making state changes from pure to mixed, that is, it decides with a certain degree of probability that it will take action that believes that the cat is alive or dead. Basically, this transformation is for the "machine scientist" to choose one of the available actions, taking $|a_1\rangle$ with probability $p_1$ (believing that the cat is alive) and taking $|a_2\rangle$ with probability $p_2$ (believing that the cat is dead) (9).

qDT determines the state of the Schrödinger's cat (29a) and calculates the observation of the Schrödinger's cat at different times (29b, $d'_{t,t-1}$ can be calculated by (26)).

$$q'_t = qDT(F, T) = \begin{cases} 0, & \text{"Machine scientists" believe with a probability } p_1 \text{ that the Schrödinger's cat is alive)} \\ 1, & \text{"Machine scientists" believe with a probability } p_2 \text{ that the Schrödinger's cat is dead)} \end{cases} \tag{29a}$$

$$x'_t = \begin{cases} x'_{t-1} + 1, & \text{if } q'_t = 0) \\ x'_{t-1} - 1, & \text{if } q'_t = 1) \end{cases} \tag{29b}$$

As shown in Figure 7, "Machine scientists" can obtain a "satisfactory" $qDT_{tree}$ (30) by using the expected value as a fitness function (13). For the "Machine scientists", this qDT provides four strategies $\{S_1, S_2, S_3, S_4\}$, and the "Machine scientists" can randomly choose a strategy from the four and apply this strategy that'll guide the "Machine scientists" in choosing which action to take (with a certain probability). If strategy $S_1$ is chosen, then the "Machine scientists" is 100% sure that the Schrödinger's cat is alive; if strategy $S_2$ is chosen, the "Machine scientists "is 100% sure that the Schrödinger's cat is always dead; if strategies $S_3$ or $S_4$ are chosen, the "Machine scientists" believes that the Schrödinger's cat is alive with 55% probability or the "Machine scientists "believes that the Schrödinger's cat is dead with 45% probability

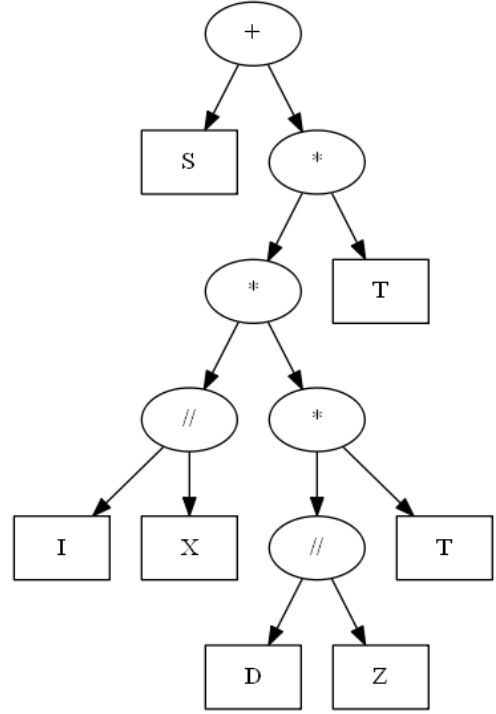

**Figure 7** qDT for Schrödinger's cat.

$$\text{qDT}_{\text{tree}} = \left(S + \left(\left((I//X) * ((D//Z) * T)\right) * T\right)\right) \tag{30}$$

- $S_1 = (S + (I * (Z * T)) * T) \rightarrow |a_1\rangle\langle a_1|$

- $S_2 = (S + (X * (D * T)) * T) \rightarrow |a_2\rangle\langle a_2|$

- $S_3 = (S + (I * (D * T)) * T) \rightarrow 0.55|a_1\rangle\langle a_1| + 0.45|a_2\rangle\langle a_2|$

- $S_4 = (S + (X * (Z * T)) * T) \rightarrow 0.55|a_1\rangle\langle a_1| + 0.45|a_2\rangle\langle a_2|$

$\text{xFT}_{\text{tree}}$ has accurately obtained an absolute distance of 1 between the two observation points of Schrödinger's cat. By combining $\text{qDT}_{\text{tree}}$ and $\text{xFT}_{\text{tree}}$, "machine scientists" can "reconstruct" the "trajectories" of Schrödinger's cat observations with 100% accuracy (Figure 8);, But "machine scientists" can only make 50/50 probability predictions about the future state of Schrödinger's cat (31). In other words, without applying the Schrödinger equation at all, let alone inventing the concept of the wave function, the "machine scientist" independently "discovered" the Bonn rule (cats are alive or dead with a 50/50 probability).

$$q'_{t+1} = \begin{cases} 0, \text{If strategy } S_1 \text{ selected or } S_3|S_4 \text{ selected and action } a_1 \text{ taken} \\ 1, \text{If strategy } S_2 \text{ selected or } S_3|S_4 \text{ selected and action } a_2 \text{ taken} \end{cases} \tag{31}$$

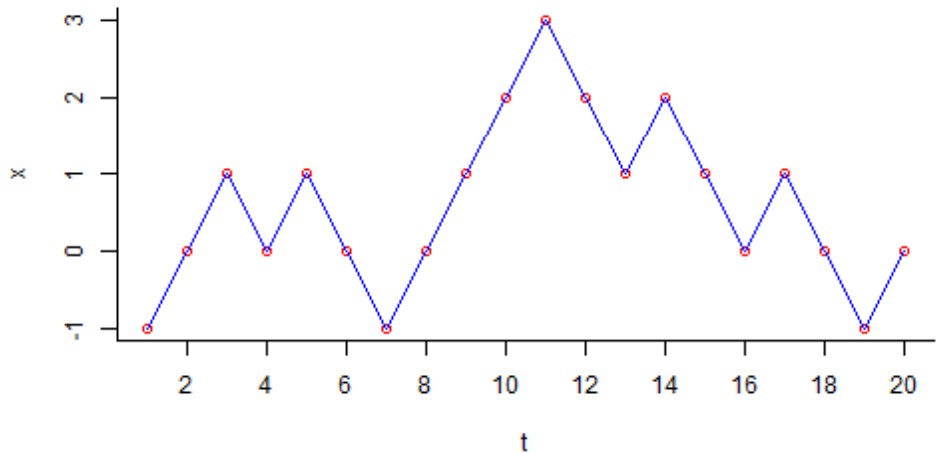

**Figure 8** "path" of Schrödinger's cat (blue line is the observed one, and the red dots are the computed ones)

## Discussion

Our knowledge of nature is derived from the observer's direct observation of nature, but knowledge itself is not objective reality. As Ludwig Boltzmann [33] pointed out 100 years ago:

*"The brain makes pictures of the world, which are useful representations of experiences …. I called theory a purely mental inner picture…. The evolution of themes and ideas take place in successive jump, just as biological evolution did. The pictures which evolved in the brain tended to perfection in the course of the centuries according to the same rules as laid down in Darwin's theory. Thus, they developed slowly as representations of experience…the sciences forward called descriptive began to triumph when Darwin's hypothesis allowed them not only to describe but also to explain biological forms and phenomenon. Almost at the same time, physics oddly enough took a turn in the opposite direction…as Hertz puts it rather characteristically, the task is merely to represent directly observed phenomena in bare equations, without the colourful wrappings of hypotheses that our imagination lends them.…The differential equations of mathematico-physical phenomenology are evidently nothing but rules for forming and combining numbers and geometrical concepts, and these in turn are nothing but mental pictures from which appearances can be predicted."*

The current mainstream scientific discovery is to explain natural phenomena by defining some basic concepts (force, field, wave function, etc.), with the aim of finding rigorous mathematical models to accurately describe nature. David Hilbert is best known for proposing the 23 most important mathematical problems (the Hilbert problem) in 1900 at the International Mathematical Congress in Paris. Hilbert firmly believes that there exists an objective world independent of us, and that our subjective consciousness by studying the objective world will eventually find a final theory described by beautiful mathematical models. In a speech in 1930, Hilbert confidently declared: "We must know, we shall know". Hilbert's dream was soon shattered. Gödel's incompleteness theorem and Turing's halting problem hint at the impossibility of the final theory (an observer who is strictly confined to a closed system is unlikely to have complete and valuable information about the closed system itself).

The world in which we live is a real world made of time and space; the introduction of axiomatic probabilities actually introduces a virtual world of possibilities in our lives. In this way, probability draws a boundary between the real world and the virtual world, and the problem is that we don't know where the exact boundary between the real and virtual worlds is. The real world is an objective world; the virtual world is a subjective world. We believe that theory is the unity of the objective world (observed data) and the subjective world (the experience of the observer). Our proposed quantum-like machine learning algorithm is to build a theory about the objective world by learning the observed data, but the theory built by "machine scientists" is constrained by our subjective experience (knowledge) and can only be continuously improved through continuous playing games with nature.

Unlike the traditional way of constructing theories with rigorous mathematical structures, our approach emphasizes on machine learning, where "machine scientists" learn observed data by applying evolutionary algorithms and construct theories to describe, predict, and understand nature. "Machine scientists" derive useful information from measured data and construct scientific theories from this. In other words, theory itself is just to obtain valuable information of nature. A scientific theory should be the unity of objective existence (nature) and subjective beliefs (observer), and be improved upon through continuous evolution.

## Declarations

**Conflicts of interest/Competing interests -** The authors declare no competing interests.
**Availability of data and material -** The data that support the findings of this study are available on request.
**Authors' contributions -** All authors conducted the research and contributed to the development of the model. H.X. contributed as an expert in quantum theory and non-linear science. L.X. contributed to the research from the aspects of machine learning, decision theory and wrote the code. L.X. and K.X. wrote this manuscript and did data analysis. All authors reviewed the manuscript.